\documentclass[runningheads]{llncs}

\usepackage{eccv}

\usepackage{eccvabbrv}

\usepackage{graphicx}
\usepackage{booktabs}
\usepackage{float}
\usepackage{wrapfig}
\usepackage{xcolor}

\usepackage[accsupp]{axessibility}  

\usepackage{hyperref}

\usepackage{orcidlink}

\begin{document}

\title{Learning to Forecast Crop Growth\\ from Earth Observation Data}

\author{
Dominik Senti\inst{1}\orcidlink{0009-0001-5648-0192} \and
Mehmet Ozgur Turkoglu\inst{1}\orcidlink{0000-0003-1446-2778} \and
Michele Volpi\inst{2}\orcidlink{0000-0003-2771-0750}\and
Helge Aasen\inst{1}\orcidlink{0000-0003-4343-0476}}

\authorrunning{D. Senti et al.}

\institute{Earth Observation of Agroecosystems Team, Agroscope, Switzerland \and {Swiss Data Science Center, ETH Zurich and EPFL, Switzerland} \\
\email{dominik.senti@agroscope.admin.ch, mehmet.tuerkoglu@agroscope.admin.ch, michele.volpi@sdsc.ethz.ch,
helge.aasen@agroscope.admin.ch}
}
\maketitle
\begin{abstract}

Forecasting crop growth across agricultural landscapes is important for improving the productivity, resilience, and operational management of farming systems. In this work, we investigate whether Earth observation time series and meteorological drivers can be used to predict future canopy development at country-scale. 
We focus on winter wheat and formulate crop growth prediction as forecasting future leaf area index (LAI) trajectories beyond the last available Sentinel-2 observation. 
We evaluate this task on a multi-year dataset which spans the entire country of Switzerland, containing over 20 million pixel-level Sentinel-2-derived LAI time series paired with meteorological variables.
Because cloud cover and revisit gaps leave LAI supervision sparse, models fit the few valid (cloud-free) LAI observations yet oscillate implausibly between them, producing trajectories no real canopy could follow.
We introduce a lightweight unimodal shape regulariser which improves trajectory plausibility with negligible loss in accuracy.
We compare deep learning sequence-to-sequence (Seq2Seq) models with classic machine learning baselines and show that Seq2Seq models generalise well across years, achieving $\text{R}^2 \text{ above }0.8$ and consistently outperforming conventional approaches. 
 Together, these results demonstrate that remote sensing and weather-driven sequence modelling can learn crop growth dynamics at landscape-scale.

\keywords{Crop Growth \and Leaf Area Index \and Forecasting \and Remote Sensing
\and Time Series \and RNN \and Transformer}
\end{abstract}

\section{Introduction}
\label{sec:intro}

Agriculture, occupying approximately one-third of the Earth's %
ice-free land surface, is pivotal for food production~\cite{foley_solutions_2011} but is threatened by climate change~\cite{rezaei_climate_2023}.
Moreover, it contributes significantly to biodiversity loss \cite{schmidhuber_global_2007,godfray_food_2010,rezaei_climate_2023,campbell_agriculture_2017,gornall_2010_ImplicationsClimateChange}, soil degradation \cite{bindraban_assessing_2012}, and environmental pollution \cite{pingali_green_2012,laborde_agricultural_2021}.
In this context, understanding growth processes is key to making agriculture more productive and resilient.

A central variable in this regard is the green leaf area index (LAI), the one-sided green leaf area per unit ground area. It is a recognised essential climate variable and a key control on biosphere--atmosphere mass and energy exchange~\cite{asner_2003_GlobalSynthesisLeaf,wmo2011,mason_2010_ImplementationPlanGlobal}. LAI tracks the crop growing cycle across a season~\cite{gitelson_relationships_2014}, governs canopy light interception, and scales with above-ground biomass. It is therefore widely used to monitor crop productivity, crop stress and health, biomass, phenology, and nutrient supply~\cite{gitelson2003,mulla2013,huang2015,chen2018,hashimoto2022,delloye2018}. For winter wheat in particular, grain yield depends on intercepted light and accumulated biomass, both well approximated by LAI, making LAI predictive of final yield~\cite{rose2017,gross2023}.

Crop growth has traditionally been studied at the experimental scale: in indoor or agronomic experiments, often carried out in small plots or with a limited number of experimental fields, that isolate individual drivers under a restricted range of conditions. The drivers of growth, however, can differ in importance across scales and environments~\cite{tang_2016_EmergingOpportunitiesChallengesa,poorter_2016_PamperedPesteredOutside}, so findings from these settings do not necessarily transfer to the landscape-scale, the heterogeneous, real-world conditions across which crops are actually grown and managed. Capturing growth at this scale demands observations with matching spatial and temporal coverage.

\begin{figure}
\vspace{-10mm}
    \centering
    \includegraphics[width=0.95\linewidth]{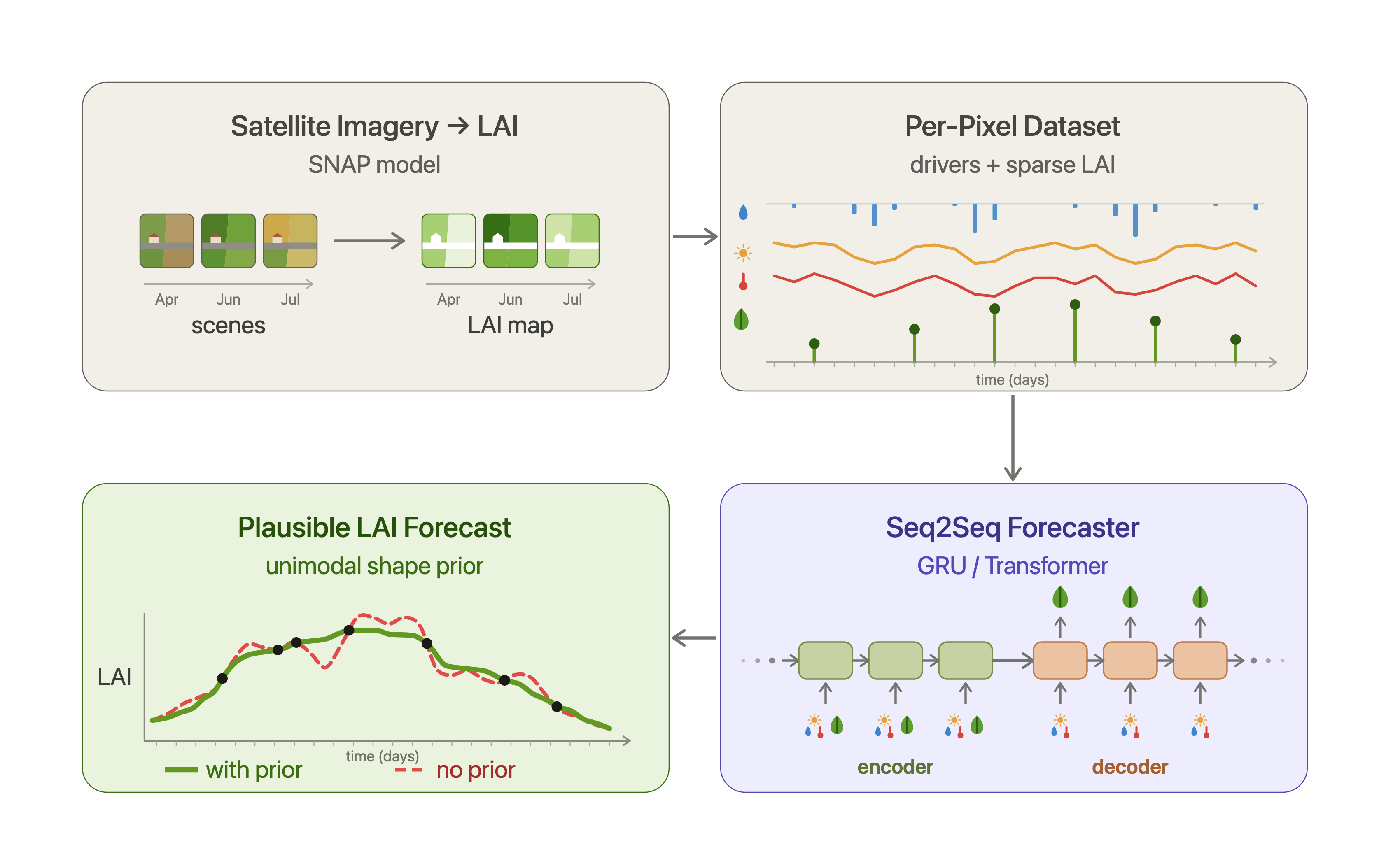}
    \caption{\textbf{Overview of the proposed framework.}
\textbf{(1) Imagery$\rightarrow$LAI:} Sentinel-2 (cloud free) scenes are inverted to per-pixel LAI by the SNAP processor. \textbf{(2) Dataset:} sparse LAI is paired with dense daily meteorological drivers into a country-scale, per-pixel dataset.
\textbf{(3) Forecaster:} a sequence-to-sequence GRU/transformer predicts LAI over the horizon from the run-in history (sparse LAI observations and dense weather) together with the future drivers.
\textbf{(4) Plausible forecast:} on sparse targets the model oscillates implausibly (no prior), while our unimodal shape regulariser restores a plausible forecast (with prior), showing a single peak per season.}
    \label{fig:framework}
\end{figure}

Satellite constellations such as Sentinel-2 from the European Space Agency (ESA) acquire Earth imagery every few days at 10\,m resolution, capturing crop growth dynamics across large regions and entire seasons~\cite{weiss_remote_2020,atzberger2013}. In parallel, modern deep sequence models, including Recurrent Neural Networks (RNNs)~\cite{elman1990,werbos1990} and transformers~\cite{vaswani_2017_AttentionAllYou}, have proven highly effective for multivariate time-series modelling and long-horizon, multi-step forecasting~\cite{nie_time_2023,liu2024arxiv,das2024decoder,lim2021ijforcasting}. Together, these advances raise a natural question: can crop growth dynamics be learned directly from Earth observation (EO) and weather data?

We take a first step towards answering this question by formulating crop growth prediction as forecasting the future leaf area index trajectory of winter wheat beyond the last available Sentinel-2 observation. Given a history of satellite-derived LAI observations and meteorological drivers, we train deep sequence-to-sequence models to predict daily LAI into the future. Unlike existing Earth observation methods that estimate crop state from contemporaneous imagery, our approach extrapolates future canopy development from its observed history and expected environmental conditions. It therefore requires learning the temporal dynamics of crop growth rather than a static mapping between imagery and crop state.

We consider forecast horizons of up to 32 days, matching the approximate monthly range increasingly supported by modern weather forecasting systems~\cite{vonich2026}. This lead time is also operationally relevant for agricultural interventions such as nitrogen top-dressing and fungicide application~\cite{zadoks1974}, as well as for updating monthly crop-condition assessments~\cite{beckerreshef2020geoglam}.

Learning these dynamics from satellite observations introduces a fundamental difficulty. Cloud cover and satellite revisit intervals leave LAI supervision sparse and irregular, although the model must produce a dense daily trajectory. When trained only on the available cloud-free observations, flexible forecasting models can match the sparse targets while oscillating implausibly between them, producing trajectories that no real winter-wheat canopy could follow. We address this issue with a differentiable unimodal shape regulariser that encodes the characteristic seasonal rise, peak, and decline of green LAI. The regulariser acts directly on the predicted trajectory, requires no additional LAI observations, and can be applied to different sequence-model architectures.

\smallskip\noindent\textbf{Contributions.}
\begin{itemize}
  \item We formulate landscape-scale crop growth forecasting as predicting future LAI trajectories beyond the last available satellite observation from historical remote sensing observations and known environmental drivers. 
  \item We introduce an ecophysiological LAI seasonal shape regulariser to allow learning under sparse supervision, encouraging biologically plausible crop growth trajectories.
  \item Using
  a multi-year, country-scale dataset comprised of 20.6 million pixel-level winter wheat time series,
  we demonstrate that crop growth dynamics can be accurately learned from Earth observations by modern machine learning methods (e.g., transformer, RNN) while respecting ecophysiological principles of crop growth.
\end{itemize}

\section{Related Work}
\label{sec:related}
\paragraph{\textbf{LAI retrieval.}}
EO provides data and tools that are well established for deriving biophysical variables that provide crucial information about the past and present of agricultural lands ~\cite{weiss_remote_2020,atzberger2013,fang2019,graf_insights_2023,perich2023}.
Recent reviews examine how such variables are calculated from optical imagery and from dense image time series~\cite{verrelst2015,kooistra2024reviews,berger2026advancing}.
The state-of-the-art to retrieve LAI from space performs inversion of leaf and canopy radiative-transfer models (RTMs): an RTM simulates spectral signatures for a given set of leaf and canopy properties and produces a look-up table from which LAI is inferred, either by matching observed spectra to the table directly or by training a model to invert it~\cite{kira_toward_2017,sun_leaf_2021,tomicek_prototyping_2021,verrelst2019,vina_comparison_2011,jacquemoud2009,kimes2000}.
The SNAP Sentinel-2 biophysical processor follows this RTM-inversion paradigm by training a neural network on PROSAIL simulations~\cite{weiss2020s2toolbox} and creates per-pixel LAI from Sentinel-2, which images the land surface with a $10\,\mathrm{m}$ ground sample distance and roughly a five-day revisit over Switzerland~\cite{claverie2018harmonized}.
The derived LAI product has been independently validated~\cite{djamai_comparison_2018,kganyago_validation_2020} and is considered state-of-the-art.
Still, crucially, it only provides LAI on cloud-free days, producing time series with unevenly spaced observations, making learning dynamics a complex process requiring special treatment.

\paragraph{\textbf{Deep learning for time series forecasting.}}
Forecasting future values of a sequence from past observation is a mature subfield, developed largely outside EO, on tasks such as finance, energy, and weather forecasting.
Before deep learning, classical (autoregressive) statistical models such as ARIMA dominated the task~\cite{hyndman2018forecasting}.
As recent surveys document, the dominant approaches have since progressed to deep architectures, from recurrent encoder--decoders~\cite{sutskever2014,cho2014learning} to attention- and transformer-based models~\cite{lim_time-series_2021,sezer_financial_2020,wen2023transformers,kim2025comprehensive}, the latter increasingly tailored to long-horizon forecasting~\cite{zhou_informer_2021,nie_time_2023}.
More recently, the field has moved towards pretrained foundation models that forecast previously unseen time-series in a zero-shot setup \cite{ansari2024chronos,hoo2025tables,das2024decoder,kottapalli2025foundation,wang2026deep}.
This literature, however, largely assumes densely and regularly sampled inputs~\cite{schirmer2022modeling,yalavarthi2024grafiti}, whereas our problem is defined by sparse supervision, reviewed in~\cite{shukla2020survey}.

\paragraph{\textbf{Deep learning on satellite image time series.}}
The dominant use of multi-temporal satellite data in agricultural applications is classification (or segmentation, alternatively), i.e., labelling each pixel or parcel with a crop type class or phenological state~\cite{ferchichi2022forecasting}.
Recurrent networks were early workhorses for this, with LSTM~\cite{hochreiter1997} and GRU~\cite{cho2014learning} encoders applied to optical and synthetic aperture radar sequences~\cite{ndikumana_deep_2018,zhou_long-short-term-memory-based_2019}, and convolutional--recurrent models such as ConvSTAR exploit multi-scale label hierarchies~\cite{turkoglu_crop_2021}.
More recently, self-attention~\cite{vaswani_2017_AttentionAllYou} and temporal-attention architectures have been applied to the same task~\cite{ruswurm_self-attention_2020,sainte_fare_garnot_multi-modal_2022}.

A separate line of work forecasts vegetation indices forward in time~\cite{ferchichi2022forecasting}, though much of it targets a single value such as end-of-season crop yield. We, however, want to focus on extrapolating a full trajectory.
In this subdiscipline, we can split research into two general directions: image-wise forecasting and location-wise forecasting (pixel- or field-scale).
The main difference here is that the location-wise forecasting does forecasting for every location independently, while the image-wise forecasting uses spatial context.
The EarthNet2021 benchmark~\cite{requenamesa2021earthnet} framed the former as a video prediction task where weather-conditioned models forecast entire future satellite images, which can easily be converted to LAI or NDVI (Normalized Difference Vegetation Index) maps. 
Architectures which aim to solve this task are ConvLSTMs~\cite{diaconu2022understanding,robin2022learning,kladny2024enhanced,ahmad2023machine,ma2022forecasting}, the multi-modal Contextformer~\cite{benson2023multimodal}, and latent-diffusion models~\cite{zhao2025vegediff}.
The second group of per-pixel or per-field forecasters uses recurrent models to predict NDVI or vegetation-health series and is also often weather-conditioned~\cite{reddy2018prediction,lees2022deep,vasilakos2022lstm}.
It includes recent work on high-resolution Sentinel-2 and PlanetScope data~\cite{farbo2024forecasting,marsetic2024forecasting}.
~\cite{iele2026probabilistic} forecasts field-level NDVI directly from sparse, irregular clear-sky Sentinel-2 acquisitions and weather on the GreenEarthNet cubes.
Most of the previously mentioned work focusses on NDVI.
A few studies instead target LAI, forecasting its trajectory from climate drivers with attention-augmented LSTMs~\cite{xiong2024predicting} or from multitemporal imagery under heat-wave stress~\cite{gobbi2019high}.
Summarising, these methods forecast a continuous vegetation variable as a time series; unlike ours, they target NDVI or coarse spatial resolution LAI ($\geq300$m), whereas we aim to forecast high-resolution LAI (10m). In this contribution we limit ourselves to pixel-scale forecasting.

\section{Crop Growth Forecasting: Setting and Formulation}
\label{sec:context}

\subsection{Development of LAI Throughout the Growing Season}
\label{subsec:lai}

Leaf area index (LAI) has a seasonal unimodal growth pattern that is mainly driven by weather \cite{porter_temperatures_1999,merz2022}, when nutrient limitations can be ruled out as in a high-intensity agricultural regime such as Switzerland.
The green LAI growing pattern is defined by ecophysiological processes:
Early-season growth rates and the peak LAI value correlate with the crop's response to temperature, water, and nutrient availability~\cite{aasenroth2022,sjulgard2024,roth2024}.
At finer temporal scales, early-season wheat leaf growth closely relates to incoming radiation, temperature and precipitation ~\cite{merz2022}.
The timing and pace of senescence are likewise linked to temperature and water availability \cite{anderegg_2020_SpectralVegetationIndices, anderegg_2021_TemporalTrendsCanopy, anderegg_2023_ThermalImagingCan}.
Wheat development is largely governed by accumulated thermal time~\cite{porter_temperatures_1999,mcmaster_growing_1997}, modulated by photoperiod and vernalisation~\cite{slafer_sensitivity_1994}.

Winter wheat is sown towards the end of one calendar year and harvested in the next, so this rise--peak--decline plays out over a single growing season spanning roughly early spring to late summer of the harvest year.
The LAI trajectory is thus a single-peaked rise, plateau, and decline whose rates and timing reflect how the crop responds to its environment.
This trajectory is the ecophysiological regularity our model is meant to respect and enforce.
\subsection{Problem Formulation}
\label{sec:formulation}

We formulate the problem as follows. We consider a winter-wheat pixel location $p$ at daily resolution over a growing season. On each day $t$, a feature vector $x_{p,t}$ is available, containing the daily meteorological drivers of growth together with static location metadata, namely latitude and longitude, and a day-of-year encoding. In contrast, the LAI target $y_{p,t}$ is available only on a subset $\mathcal{V}_p$ of \emph{valid days}, corresponding to cloud-free Sentinel-2 acquisitions from which an LAI value can be retrieved. These valid observations provide the only LAI supervision available to the model.

We fix a run-in length $R_{\mathrm{in}} = 90$ days, which in the beginning of the season is long enough to include the sowing date (cf.\ Sec.~\ref{subsec:lai}), and forecast horizons of up to $H = 32$ days. These are the values used throughout our experiments.
Around a reference day $t_0$ the timeline is split into a run-in window $\mathcal{W}_{\mathrm{in}}$ and a prediction window $\mathcal{W}_{\mathrm{pred}}$,
\begin{equation}
  \mathcal{W}_{\mathrm{in}} = \{\, t_0 - R_{\mathrm{in}} + 1,\, \dots,\, t_0 \,\},
  \qquad
  \mathcal{W}_{\mathrm{pred}} = \{\, t_0 + 1,\, \dots,\, t_0 + H \,\}.
  \label{eq:windows}
\end{equation}
Over the run-in window the model sees the full history, both the features $x_{p,t}$ and any valid LAI observation $y_{p,t}$, while over the prediction window it sees only the features and must forecast LAI.
The model is a learned map $f$ that, from the run-in history and the future drivers, predicts an LAI value for every day of the horizon,
\begin{equation}
  \hat{y}_{p,t}
  = f\!\Bigl(
      \{ (x_{p,i},\, y_{p,i}) \}_{i \in \mathcal{W}_{\mathrm{in}}},\;
      \{ x_{p,i} \}_{i \in \mathcal{W}_{\mathrm{pred}}}
    \Bigr),
  \qquad t \in \mathcal{W}_{\mathrm{pred}}.
  \label{eq:forecast}
\end{equation}
The future drivers $\{ x_{p,i} \}_{i \in \mathcal{W}_{\mathrm{pred}}}$ are taken from observed weather data; in operational deployment they would be replaced with meteorological forecasts. No satellite-derived LAI enters the prediction window; observations there are used only for supervised training and evaluation.

\subsection{Dataset}
\label{subsec:dataset}

We assemble a country-scale dataset of winter wheat LAI time series for Switzerland, pairing satellite-derived LAI with the meteorological drivers of growth over five growing seasons (2021--2025).

\paragraph{\textbf{Field locations.}}
Winter wheat fields are taken from the agricultural parcel datasets published by the Swiss federal and cantonal authorities~\cite{geodienste_nutzungsflaechen}, where each parcel carries a crop-type code. This data is also part of the SwissCrop25 national crop-mapping benchmark~\cite{lauber2026swisscrop}, which is publicly available on Hugging Face.
We rasterise the winter wheat polygons onto a $10\,\mathrm{m}$ grid and keep only pixels fully contained within a field, excluding boundary pixels that may mix in roads, trees, or adjacent fields.

\paragraph{\textbf{Meteorological drivers.}}
Daily weather is obtained from the Swiss national weather service MeteoSwiss, and represents gridded daily weather data~\cite{meteoswiss_griddata} at $1\,\mathrm{km}$ resolution: minimum, mean, and maximum temperature, relative sunshine duration, and precipitation.
For each pixel we also compute cumulative aggregates of these variables from the 1st of January of the sowing year, so run-in windows of any length are covered.
Temperature enters as $\max(T,\,0\,^{\circ}\mathrm{C})$, since wheat growth and development effectively cease below $0\,^{\circ}\mathrm{C}$~\cite{porter_temperatures_1999,mcmaster_growing_1997}; clipping at zero lets sub-zero days simply not contribute to the cumulative aggregate rather than subtracting from previously accumulated temperature.
Day-of-year and days-since-start (counted from the 1st of January of the sowing year), together with pixel latitude and longitude, are included as light spatio-temporal metadata.

\paragraph{\textbf{LAI extraction and cleaning.}}
Sentinel-2 scenes are cleaned with a two-stage mask: a U-Net cloud-and-shadow segmentation model~\cite{benson2023multimodal} combined with the Sentinel-2 Scene Classification Layer, retaining only pixels flagged clear by both.
In case of double acquisitions on the same day (due to overlapping satellite orbits), we keep the scene with the lowest post-masking cloud fraction, and derive per-pixel LAI time series on the surviving cloud-free days with the SNAP Sentinel-2 biophysical processor (see Sec. \ref{sec:related}, LAI retrieval).
Two further checks enforce valid winter wheat LAI values: we discard any value above LAI $8$, since these are highly implausible for Swiss winter wheat and usually result from residual cloud contamination~\cite{graf_insights_2023}, and exclude parcels falsely labelled as winter wheat by removing time series with implausible early- or late-season dynamics by requiring $\mathrm{LAI}<3$ on days $50$ and $210$ of the harvest year.
Together these thresholds remove fewer than $2\%$ of pixels.
The final dataset comprises approximately $20.6$ million pixel-level LAI time series across five seasons covering Switzerland.

\paragraph{\textbf{Spatial split.}}
To measure geographic generalisation we split the pixel locations into disjoint stripes, holding out four evenly spread validation stripes (about $15\%$ of the patches) and training on the rest.
The stripes are interleaved four times west to east so the validation set spans the full climatic gradient of Switzerland but with reduced spatial correlation across sets.
App.~\ref{sec:appendix-split} shows the resulting split.

\section{Method}
\label{sec:method}

We cast LAI forecasting as a sequence-to-sequence learning problem. For each pixel $p$, the encoder processes the run-in window, comprising the meteorological drivers $x_{p,t}$ and the sparse LAI observations available on days $t \in \mathcal{V}_p$, and maps this history to a latent representation of the current crop state. Conditioned on this representation and the meteorological drivers over the prediction window $\mathcal{W}_{\mathrm{pred}}$, the decoder produces a daily LAI forecast $\hat{y}_{p,t}$ for each step of the forecast horizon. The run-in window should be sufficiently long to capture the recent crop trajectory and partially account for latent environmental conditions, such as soil moisture, that are not directly observed; we discuss this further in Appendix~\ref{sec:appendix-soil}.

We consider two encoder--decoder architectures. The first is a recurrent sequence-to-sequence model with a bidirectional GRU encoder and a unidirectional GRU decoder~\cite{cho2014learning}. The decoder generates the forecast sequentially, one timestep at a time. However, previous LAI predictions are not fed back into later decoding steps, preventing the accumulation of autoregressive rollout errors. The second is a transformer encoder--decoder~\cite{vaswani_2017_AttentionAllYou}, which models temporal dependencies through self-attention and predicts the complete forecast horizon in a single non-autoregressive pass.

Because Sentinel-2-derived LAI is unavailable on cloudy or otherwise invalid acquisition days, the run-in sequence contains missing observations. On such days, the LAI input is replaced by a fixed placeholder and accompanied by a binary validity mask, allowing the encoder to distinguish missing values from genuine LAI measurements.

All models are trained using a masked mean-squared error computed only on prediction-window days for which a valid LAI target is available. For a forecast window associated with pixel $p$, the loss is

\begin{equation}
  \mathcal{L}_{\mathrm{MSE}}
  =
  \frac{1}{\lvert \mathcal{O}_p \rvert}
  \sum_{t \in \mathcal{O}_p}
  \left(
    \hat{y}_{p,t} - y_{p,t}
  \right)^2,
  \label{eq:mse}
\end{equation}

where $\mathcal{O}_p \subseteq \mathcal{W}_{\mathrm{pred}}$ denotes the subset of forecast days with valid LAI observations. The loss is averaged across the windows in each training batch. Prediction days without a valid LAI target do not contribute to the supervised loss or its gradients.

\subsection{Ecophysiological Prior via Unimodal Shape Regularisation}
\label{subsec:prior}

\begin{figure}[t]
  \centering
  \includegraphics[width=0.98\linewidth]{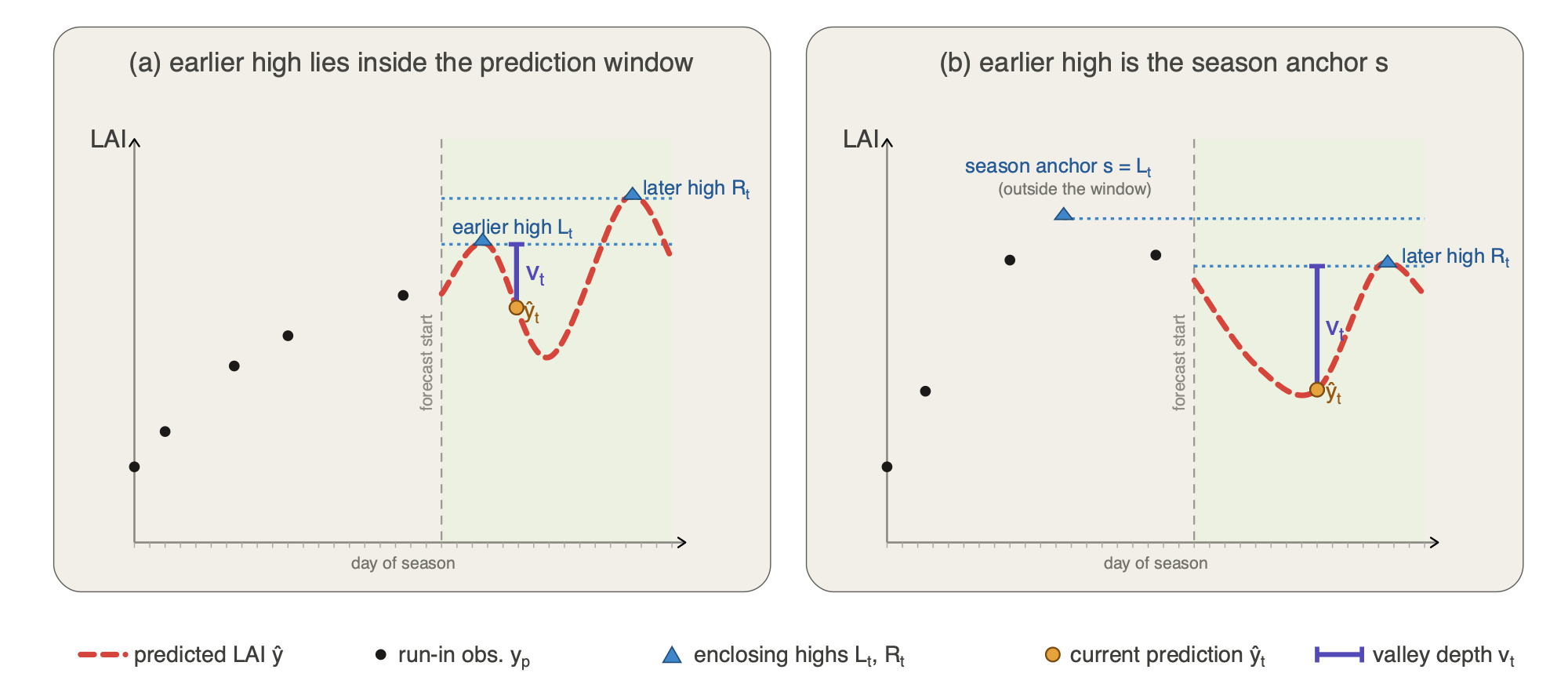}
  \caption{\textbf{The unimodal shape regulariser penalises interior valleys.}
  A prediction $\hat{y}_t$ that lies below both an earlier high $L_t$ and a later
  high $R_t$ violates unimodality. The penalty $v_t$ is the depth of this valley
  relative to the lower of the two surrounding highs.
  \textbf{(a)} The earlier high lies inside the forecast window.
  \textbf{(b)} The season anchor $s$, obtained from valid run-in observations,
  represents a peak preceding the forecast window.}
  \label{fig:valley}
\end{figure}

The masked loss constrains predictions only on days with valid LAI observations.
Consequently, models can fit the sparse targets while producing oscillatory
trajectories between them. This is inconsistent with the seasonal development of
green winter-wheat LAI, which follows a unimodal rise--peak--decline when not under nutrient or disease stress.
We therefore penalise interior valleys in the predicted trajectory
(Fig.~\ref{fig:valley}).

Because the forecast window is short relative to the full growing season, a
seasonal peak may occur before the prediction window. Let $\mathcal{S}_p$ denote
the valid run-in observations from the harvest season being forecast. We define
the season anchor as
\begin{equation}
  s = \max_{t \in \mathcal{S}_p} y_{p,t}.
  \label{eq:anchor}
\end{equation}
For each forecast step $t$, the highest values before and after $\hat{y}_t$ are
\begin{equation}
  L_t = \max\{s,\hat{y}_1,\dots,\hat{y}_{t-1}\},
  \qquad
  R_t = \max\{\hat{y}_{t+1},\dots,\hat{y}_H\}.
  \label{eq:runningmax}
\end{equation}
We take the empty maximum as $-\infty$, so the final prediction cannot be
penalised as a valley. The valley depth is
\begin{equation}
  v_t =
  \min\!\left(
    \operatorname{ReLU}(L_t-\hat{y}_t),
    \operatorname{ReLU}(R_t-\hat{y}_t)
  \right),
  \label{eq:valley}
\end{equation}

which is positive only when $\hat{y}_t$ lies below both an earlier and a later
high. The regulariser and total training objective are
\begin{equation}
  \mathcal{R}(\hat{y})
  = \frac{1}{H}\sum_{t=1}^{H} v_t,
  \qquad
  \mathcal{L}
  = \mathcal{L}_{\mathrm{MSE}}
  + \lambda \mathcal{R}(\hat{y}),
  \label{eq:reg}
\end{equation}

where $\lambda \geq 0$ controls the strength of the prior.  
In \ref{subsec:validate-prior} we show how we choose $\lambda$ to be both ecophysiologically viable while keeping model performance. 
Since
$\mathcal{R}(\hat{y})$ depends only on the dense predicted trajectory, it also
provides a training signal on days without LAI supervision. It is used only
during training and applies unchanged to both the GRU and transformer models.

\section{Experiment and Discussion}
\label{sec:experiments}

\subsection{Baselines}
\label{subsec:baseline}
To assess the benefit of explicit temporal modelling, we compare the GRU and transformer encoder--decoder models with two non-sequential baselines: a multilayer perceptron (MLP)~\cite{rumelhart1986learning} and LightGBM~\cite{ke2017lightgbm}. Both receive the same run-in and prediction-window variables as the sequence models, but flattened into a fixed-dimensional input vector. The MLP predicts the complete LAI trajectory jointly, whereas LightGBM performs one-step prediction and is applied recursively over the forecast horizon. These baselines allow us to isolate the contribution of architectures that explicitly model temporal structure.

\subsection{Experimental Setup and Model Selection}
\label{subsec:setup}

Our dataset spans the 2021--2025 seasons.
We define geographical splits as described in Sec.~\ref{subsec:dataset}.
We leave out 2023 completely, which is our test year, and train on the training set of the remaining years.
We use the left out validation set to select the best model architectures.
Due to compute constraints the metrics on the validation set are calculated on a fixed 25\% subset of the validation data, randomly defined.
Following the framing of Sec.~\ref{sec:formulation} we always use a run-in window of $R_{\mathrm{in}}=90$ days and for training a variable prediction horizon drawn uniformly at random in $H\in[1,32]$ days.
For performance evaluation, we use the 32-day prediction horizon.
We only allow prediction windows inside the wheat season (1~February--31~July of the harvest year).
The inputs are the LAI history together with the meteorological and time features defined in Sec.~\ref{subsec:dataset}.

\begin{wrapfigure}[22]{r}{0.48\textwidth}
  \includegraphics[width=0.48\textwidth]{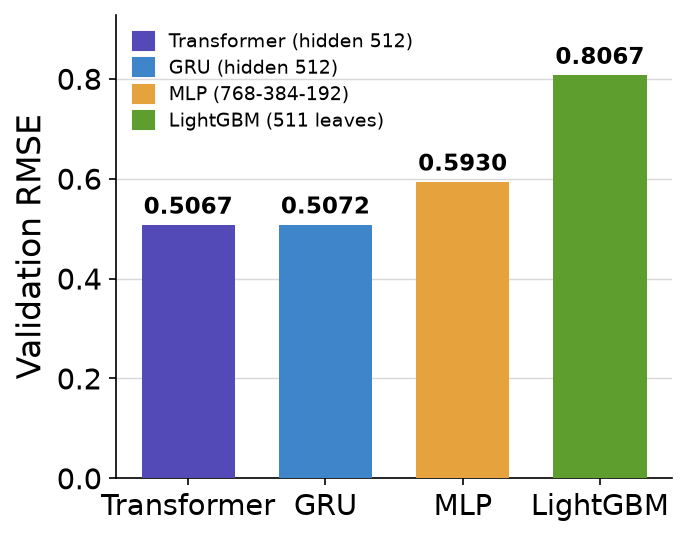}
  \caption{\textbf{Accuracy of the best model of each family.} Models trained on all seasons
  except 2023 and evaluated on the held-out geographically split validation set, measured as RMSE in
  LAI units (lower is better). The two deep encoder--decoder sequence models are
  the most accurate and essentially tied, ahead of the MLP and LightGBM
  baselines.}
  \label{fig:model-selection}
\end{wrapfigure}

In the first step, we compare the two encoder--decoder sequence models of Sec.~\ref{sec:method} (a bidirectional GRU sequence-to-sequence model and a transformer encoder--decoder) against the two non-sequential baselines (MLP and LightGBM), described in Sec.~\ref{subsec:baseline}.
All models share the identical data and evaluation protocol; optimiser and training settings are given in App.~\ref{sec:appendix-capacity}.
For each model family we sweep the model capacity and keep the size with the best validation loss. For both sequence models accuracy saturates at hidden size $512$, which we carry forward.

Comparing the best model from each family on the multi-year validation set, the two deep sequence models are the most accurate at $\approx0.51$ LAI RMSE and are essentially tied, ahead of the MLP ($0.59$) and LightGBM ($0.81$) baselines (Fig.~\ref{fig:model-selection}).

\subsection{Shape Regulariser Strength Selection}
\label{subsec:validate-prior}

\begin{figure}[t]
  \centering
  \includegraphics[width=0.49\linewidth]{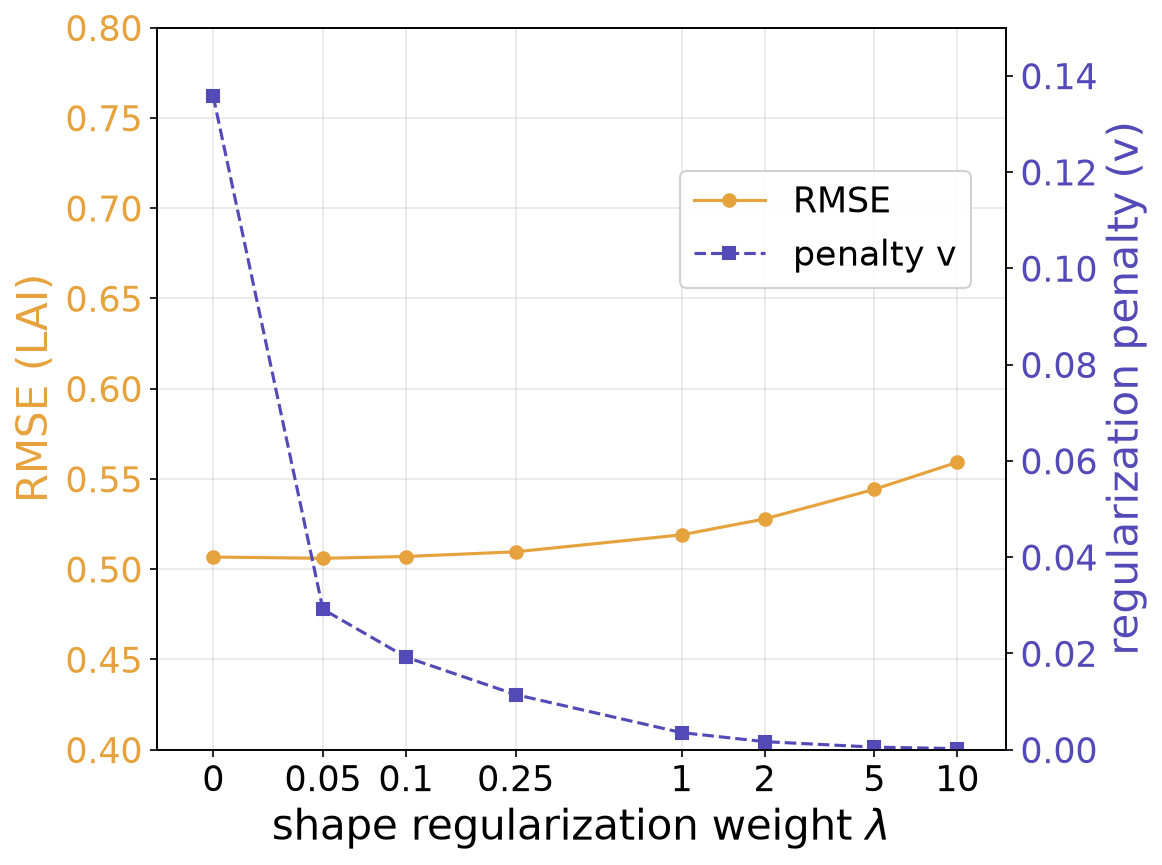}
  \hfill
  \includegraphics[width=0.49\linewidth]{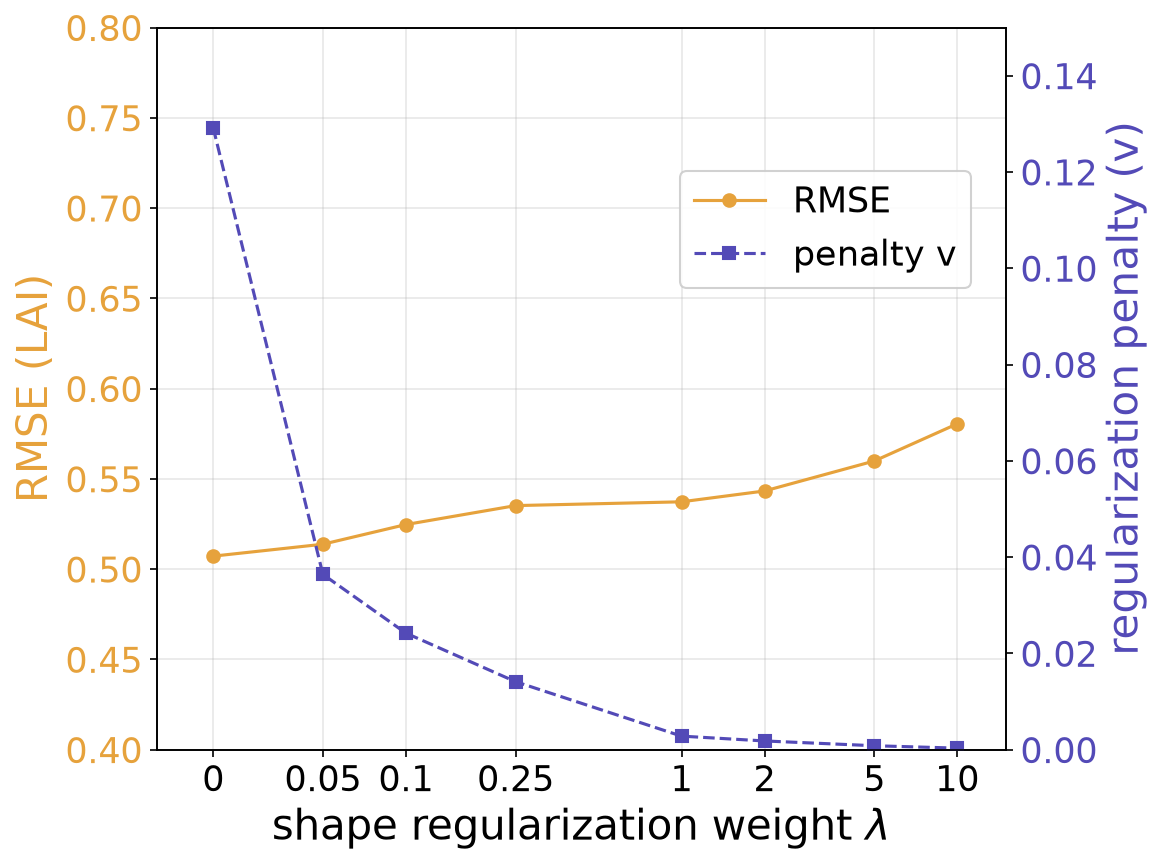}
  \caption{\textbf{Shape-regularisation trade-off for the transformer (left) and
  bidirectional GRU (right).} As the penalty weight $\lambda$ increases, the
  valley penalty $\mathcal{R}(\hat{y})$ on held-out windows (purple, right axis)
  collapses while forecast RMSE (yellow, left axis) is nearly unchanged.}
  \label{fig:lambda}
\end{figure}

\paragraph{\textbf{Plausibility versus accuracy.}}
We analyse how the validation loss behaves with increasing regularisation strength. On the two selected hidden-$512$ architectures (the GRU and the transformer), we sweep the shape-regulariser weight $\lambda$ (Eq.~\eqref{eq:reg}) over $\lambda\in\{0,0.05,0.1,0.25,1,2,5,10\}$, where $\lambda=0$ is the unregularised reference.
We measure plausibility with the valley penalty $\mathcal{R}(\hat{y})$ on the model's own predictions (lower means smoother and more strictly unimodal).
Increasing $\lambda$ lowers the valley penalty on held-out windows (Fig.~\ref{fig:lambda}, purple).
Accuracy (RMSE, yellow) stays almost flat over a wide range of $\lambda$ and increases only at the largest weights.
This trade-off is reassuring (plausibility costs almost no accuracy), but it does not make the choice of $\lambda$ unambiguous: the penalty falls smoothly and accuracy moves only slowly.
To identify a single value of $\lambda$, we next evaluate whether the predicted growth dynamics remain coupled to a known key environmental driver: temperature.

\begin{figure}[t]
  \centering
  \includegraphics[width=0.49\linewidth]{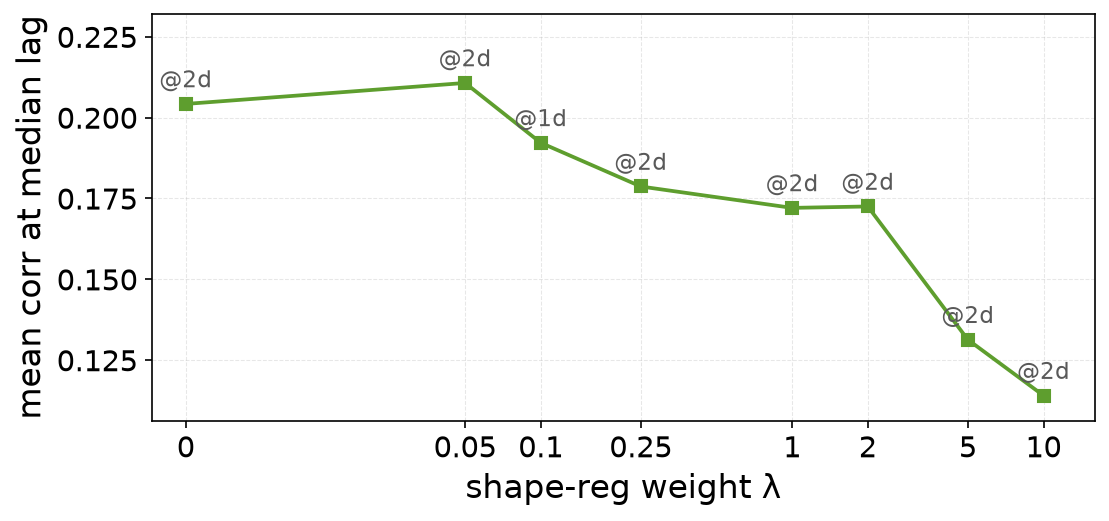}
  \hfill
  \includegraphics[width=0.49\linewidth]{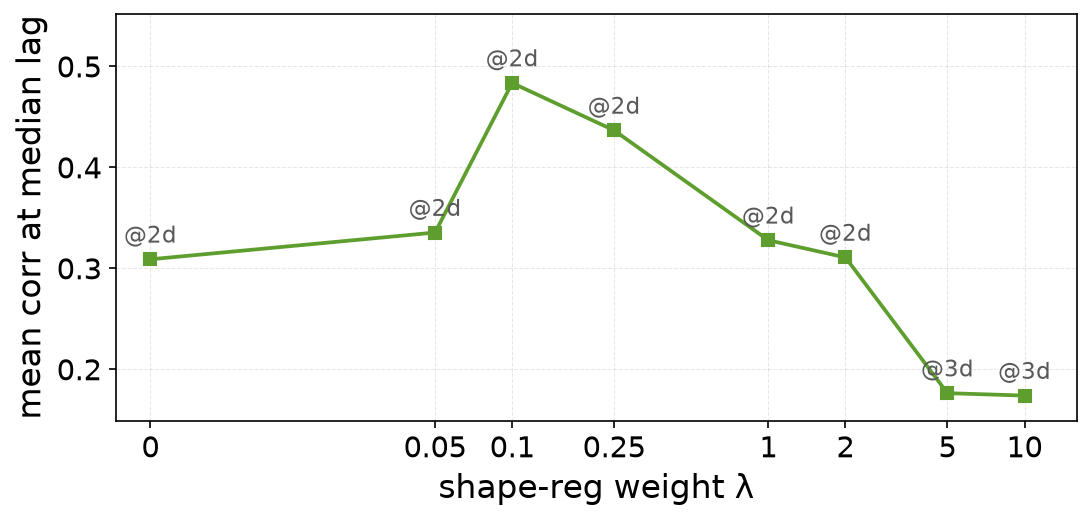}
  \caption{\textbf{Physical selection of $\lambda$ for the transformer (left) and
  GRU (right).} Strength of the correlation between predicted day-to-day LAI growth
  ($\Delta$LAI) and air temperature, read at the median response lag, plotted
  against $\lambda$; the lag is annotated above each marker (\texttt{@}$d$, in days).
  The coupling peaks at $\lambda = 0.05$ (transformer) and $\lambda = 0.1$ (GRU),
  the values we adopt. See text for the analysis and validation setup.}
  \label{fig:thermal-lambda}
\end{figure}

\paragraph{\textbf{Growth--temperature response.}}
Canopy growth is driven mainly by temperature, with leaf appearance and expansion scaling with accumulated thermal time~\cite{porter_temperatures_1999,mcmaster_growing_1997}.
A physically faithful forecast should therefore reproduce this coupling: its day-to-day growth should correlate strongly with temperature.
For each $\lambda$ we run the selected model on the validation subset and take day-to-day growth as the increments $\Delta\mathrm{LAI}_t = \hat{y}_t - \hat{y}_{t-1}$.
We correlate these increments with daily mean temperature.
We pair each increment $\Delta\mathrm{LAI}_t$ with the temperature $\ell$ days earlier, $T_{t-\ell}$, compute the Pearson correlation across the window for every offset $\ell \in \{0,\dots,14\}$, and keep the strongest correlation as that window's coupling strength.
Per-window values are averaged first by forecast start day and then across the season, giving one season-averaged coupling strength per $\lambda$.
We always use the strongest correlation over the offsets to account for delay effects, and Fig.~\ref{fig:thermal-lambda} labels each point with its median lag.
As in the \textit{plausibility versus accuracy} experiment, we restrict the analysis to forecast start days (reference days $t_0$) in DOY~60--90.
Since the validation set is inferred with a fixed $32$-day horizon, this covers predicted days 61--122, the early-spring vegetative phase when canopy growth is fast and closely tied to temperature~\cite{roth2022phenomics,roth2024}.

Fig.~\ref{fig:thermal-lambda} plots this coupling strength against $\lambda$ for both architectures.
The coupling is weak for the unregularised model, strengthens as the penalty is switched on, peaks, and weakens again for larger weights as the penalty begins to over-smooth and distort the dynamics.
We therefore select $\lambda = 0.1$ for the GRU and $\lambda = 0.05$ for the transformer: each value maximises the correlation between predicted growth and its meteorological driver while staying within the favourable accuracy--plausibility band of Fig.~\ref{fig:lambda}.
Both families behave the same way, with the implied response lag holding at a physically plausible $\sim$2 days throughout the sweep.
The season-averaged response curves at the selected $\lambda$ for both families are shown in App.~\ref{fig:thermal-curves}.

\begin{figure}[H]
  \centering
  \includegraphics[width=\linewidth]{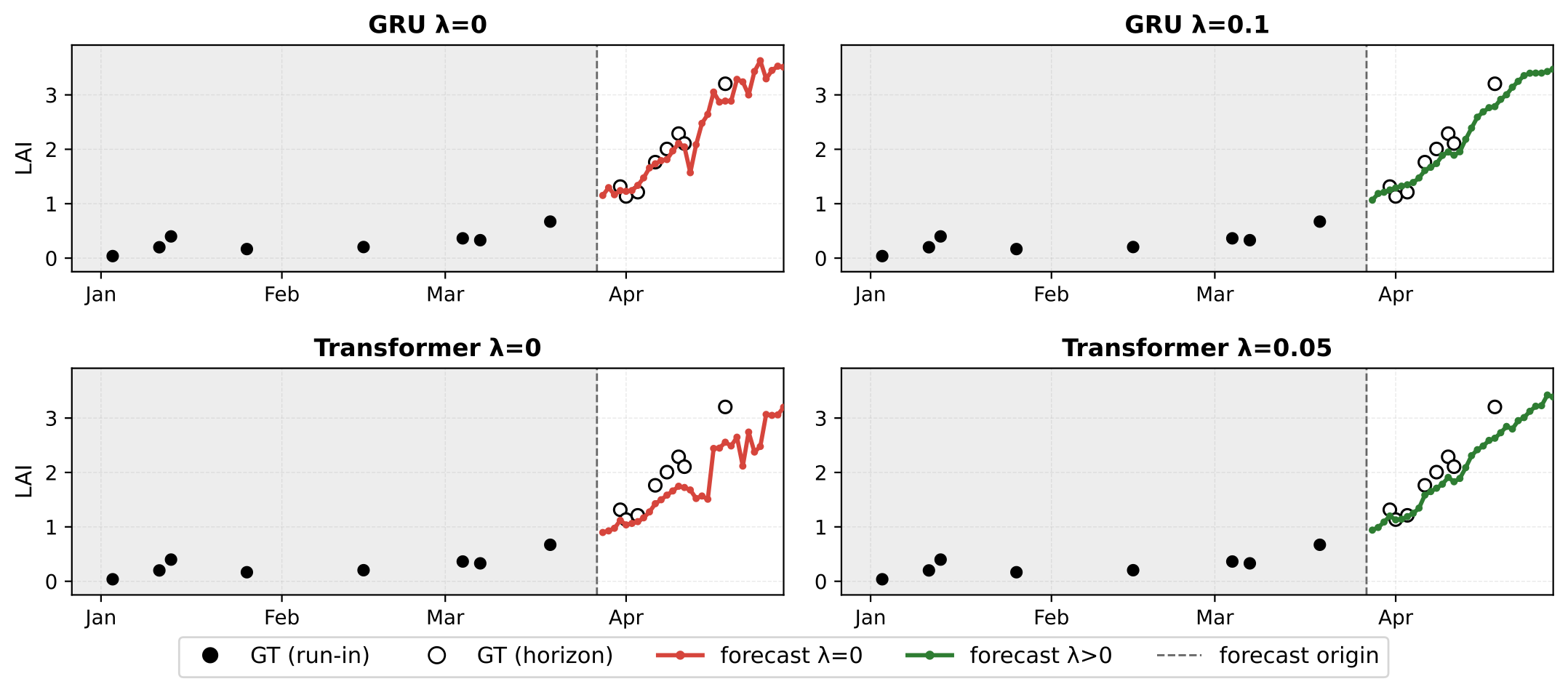}
  \caption{\textbf{The shape regulariser removes ecophysiologically impossible
  oscillations.} For one validation window (90-day run-in,
  shaded, 32-day forecast in white), each panel shows one model's predicted
  LAI trajectory against the valid run-in LAI observations and the
  forecast-horizon targets. Without the shape regulariser ($\lambda=0$) both the GRU and the transformer fit the sparse targets with jagged curves that no wheat canopy could follow. Adding the unimodal shape regulariser yields smooth, single-peaked growth.}
  \label{fig:shapereg_forecasts}
\end{figure}

Fig.~\ref{fig:shapereg_forecasts} contrasts both families' forecasts on one validation window. Without the regulariser, the GRU and transformer fit the sparse LAI observations with jagged, multi-peaked trajectories; the unimodal prior (at each family's selected $\lambda$) restores the smooth, single-peaked growth a wheat canopy actually follows.

\begin{table*}[t]
  \centering
  \caption{\textbf{Per-year LOYO forecast accuracy} for the two selected models, each season held out in turn and scored over the whole country. RMSE and MAE are in LAI units (lower is better for both), NRMSE expresses the RMSE as a percentage of the mean observed LAI (lower is better), and coefficient of determination $\text{R}^2$ (higher is better).
  The bottom row is the mean over all five folds.}
  \label{tab:per-year}
  \resizebox{\linewidth}{!}{%

  \begin{tabular}{l@{\hspace{1.5em}}cccc@{\hspace{1.5em}}cccc}
    \toprule
     & \multicolumn{4}{c}{GRU ($\lambda{=}0.1$)} & \multicolumn{4}{c}{Transformer ($\lambda{=}0.05$)} \\
    \cmidrule(lr){2-5}\cmidrule(lr){6-9}
    Test year & RMSE $\downarrow$ & NRMSE\,\% $\downarrow$ & $\text{R}^2 \uparrow$ & MAE $\downarrow$ & RMSE $\downarrow$ & NRMSE\,\% $\downarrow$ & $\text{R}^2 \uparrow$ & MAE $\downarrow$ \\
    \midrule
    2021          & 0.651 & 27.6 & 0.862 & 0.451 & 0.709 & 30.1 & 0.837 & 0.475 \\
    2022          & 0.683 & 36.8 & 0.856 & 0.465 & 0.802 & 43.2 & 0.802 & 0.512 \\
    2023          & 0.758 & 26.5 & 0.832 & 0.537 & 0.873 & 30.5 & 0.778 & 0.611 \\
    2024          & 0.867 & 38.0 & 0.725 & 0.608 & 0.795 & 34.9 & 0.769 & 0.546 \\
    2025          & 0.695 & 28.0 & 0.842 & 0.479 & 0.721 & 29.1 & 0.830 & 0.502 \\
    \midrule
    Mean & 0.731 & 31.4 & 0.823 & 0.508 & 0.780 & 33.6 & 0.803 & 0.529 \\
    \bottomrule
  \end{tabular}
  }
\end{table*}

\subsection{Leave-One-Year-Out Performance Comparison}
\label{subsec:main-results}

We compare the two selected sequence models, the GRU at $\lambda = 0.1$ and the transformer at $\lambda = 0.05$ (Sec.~\ref{subsec:validate-prior}), under a leave-one-year-out (LOYO) protocol.
Model architectures and regulariser strengths were selected on geographically held-out validation data drawn from the training seasons, without access to any held-out test season. 
We now evaluate temporal generalisation: each season in $2021$--$2025$ is held out in turn while the model is trained on the remaining four, using exactly the same training splits as in Sec.~\ref{subsec:setup}. 
The training regime is therefore unchanged; we report accuracy on the held-out season, pooled over the whole country (Table~\ref{tab:per-year}), using root-mean-square error (RMSE) and mean absolute error (MAE), both in LAI units, its scale-free normalisation NRMSE (the RMSE as a percentage of the mean observed LAI), together with the coefficient of determination $\text{R}^2$ (see App.~\ref{sec:appendix-metrics} for details).

\begin{wrapfigure}[14]{r}{0.5\textwidth}
  \centering
  \vspace{-\intextsep}
  \includegraphics[width=0.5\textwidth]{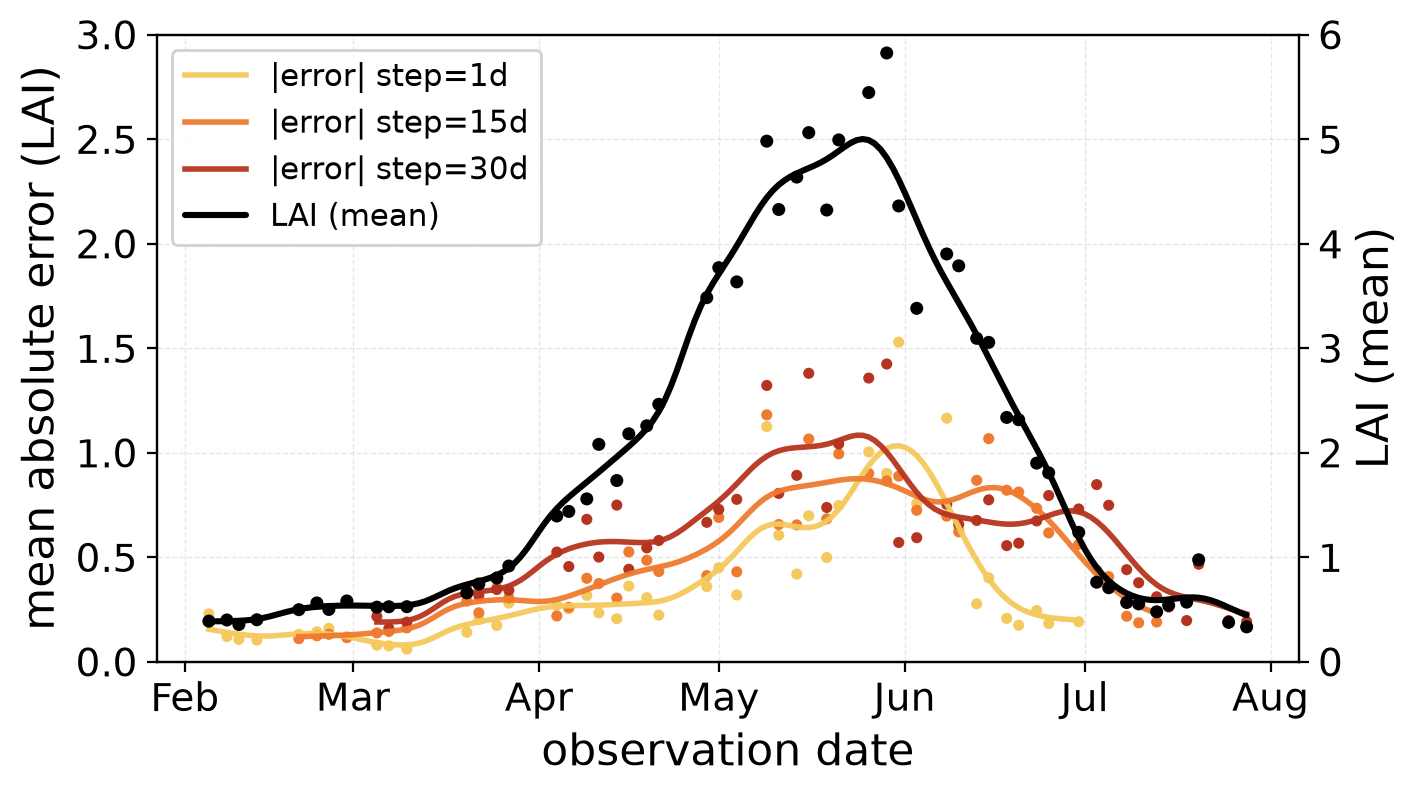}
  \caption{\textbf{Error grows in prediction horizon.} (GRU, Z\"urich region, 2022 held-out season): mean absolute error of predictions $1$/$15$/$30$-day into the future and mean LAI.}
  \label{fig:loyo-abserr-horizon}
\end{wrapfigure}

The four metrics rank the seasons quite consistently, with one notable exception for the GRU: 2023 records the lowest NRMSE of all folds ($26.5\%$) despite the second-highest RMSE. 
This decoupling is a product of the season rather than the model: 
NRMSE divides the RMSE by the mean observed LAI (Appendix, Eq.~\eqref{eq:nrmse}), and 2023 saw unusually rapid spring growth, so the canopy spent little of the season at low leaf area, there are few low-LAI observations, the mean observed LAI is correspondingly high, and the NRMSE is pulled down. The seasonal NDVI curves in App.~\ref{fig:ndvi-greenup} show this directly.

Forecast accuracy degrades only gently as the distance from the observed LAI values grows. As an illustration, Fig.~\ref{fig:loyo-abserr-horizon} follows the GRU through the 2022 held-out season over the hundred $1.28\times1.28\,\mathrm{km}$ patches nearest the Z\"urich city centre which contain winter wheat: for every Sentinel-2 observation date it plots the mean absolute forecast error at $1$-, $15$- and $30$-day prediction horizon positions, averaged over all pixel time series of those patches, together with their mean ground-truth LAI, each smoothed with a Gaussian kernel ($\sigma = 5$ days). The longer-lead forecasts carry a somewhat larger error than the short-lead ones, but all three lead times track the seasonal LAI trajectory. The transformer behaves the same way (App.~\ref{fig:loyo-abserr-horizon-tfm}).
We want to emphasize that throughout our work, we use observed weather instead of weather forecasts which would carry uncertainty. Therefore,  when forecasting into the future, our performance estimates represent an optimistic upper bound.

\section{Conclusion}
\label{sec:conclusion}

We formulated crop-growth forecasting as predicting the future LAI trajectory of winter wheat beyond the last available satellite observation from historical remote sensing imagery and meteorological drivers. To study this task at landscape-scale, we assembled a country-scale dataset for Switzerland comprising approximately 20.6 million pixel-level LAI time series across the 2021--2025 growing seasons, pairing sparse Sentinel-2-derived LAI observations with dense daily meteorological variables.
We trained deep sequence-to-sequence encoder--decoder models based on GRU and transformer architectures and compared them with non-sequential baselines. Both sequence models consistently achieved higher forecasting accuracy, showing that explicit temporal modelling is important for learning crop growth dynamics. Under leave-one-year-out evaluation, they maintained strong performance across growing seasons.
Because cloud cover and revisit gaps provide LAI supervision only on sparse, irregular days, unconstrained models can produce implausibly oscillating trajectories between observations. We therefore introduced a lightweight unimodal shape regulariser that encourages the characteristic seasonal rise and decline of winter-wheat LAI without requiring additional labels. The regulariser substantially improves trajectory plausibility while largely preserving accuracy.

Overall, our results show that crop growth dynamics can be forecast from satellite-derived observations and meteorological drivers at landscape-scale. Future work should examine additional crop species and regions, incorporate further growth drivers such as soil and management information, analyse the effect of run-in window sparsity on forecasting accuracy, and evaluate the use of uncertain meteorological forecasts in operational settings.

\clearpage
\section*{Acknowledgements}
We thank the anonymous reviewers for their constructive comments. 
We also thank Thomas Lauber (Agroscope) for his help with the HPC environment, Sélène Ledain (Agroscope) for preparing several of the data products and Achim Walter and Konrad Schindler (both ETHZ) for their valuable input.
This work was supported by the Swiss National Science Foundation (Grant No. 10002727) and Agroscope. We acknowledge access to Alps at the Swiss National Supercomputing Centre, Switzerland under Agroscope's share with the project ID go57. All funding was awarded to Helge Aasen.
Large language models were used in the preparation of this manuscript for writing assistance, language editing, and code development. All scientific content, experimental results, and conclusions were verified by the authors.

\clearpage

\bibliographystyle{splncs04}
\bibliography{main}

\clearpage

\clearpage

\appendix
\renewcommand{\thefigure}{\Alph{section}}
\renewcommand{\figurename}{App.}
\renewcommand{\theequation}{\thesection.\arabic{equation}}
\makeatletter\@addtoreset{equation}{section}\makeatother
\renewcommand{\theHsection}{appendix.\Alph{section}}
\renewcommand{\theHsubsection}{appendix.\Alph{section}.\arabic{subsection}}
\renewcommand{\theHfigure}{appendix.\arabic{figure}}
\renewcommand{\theHtable}{appendix.\arabic{table}}
\renewcommand{\theHequation}{appendix.\Alph{section}.\arabic{equation}}
\section{Spatial Split Map}
\label{sec:appendix-split}

\begin{figure}[H]
  \centering
  \includegraphics[width=0.85\linewidth]{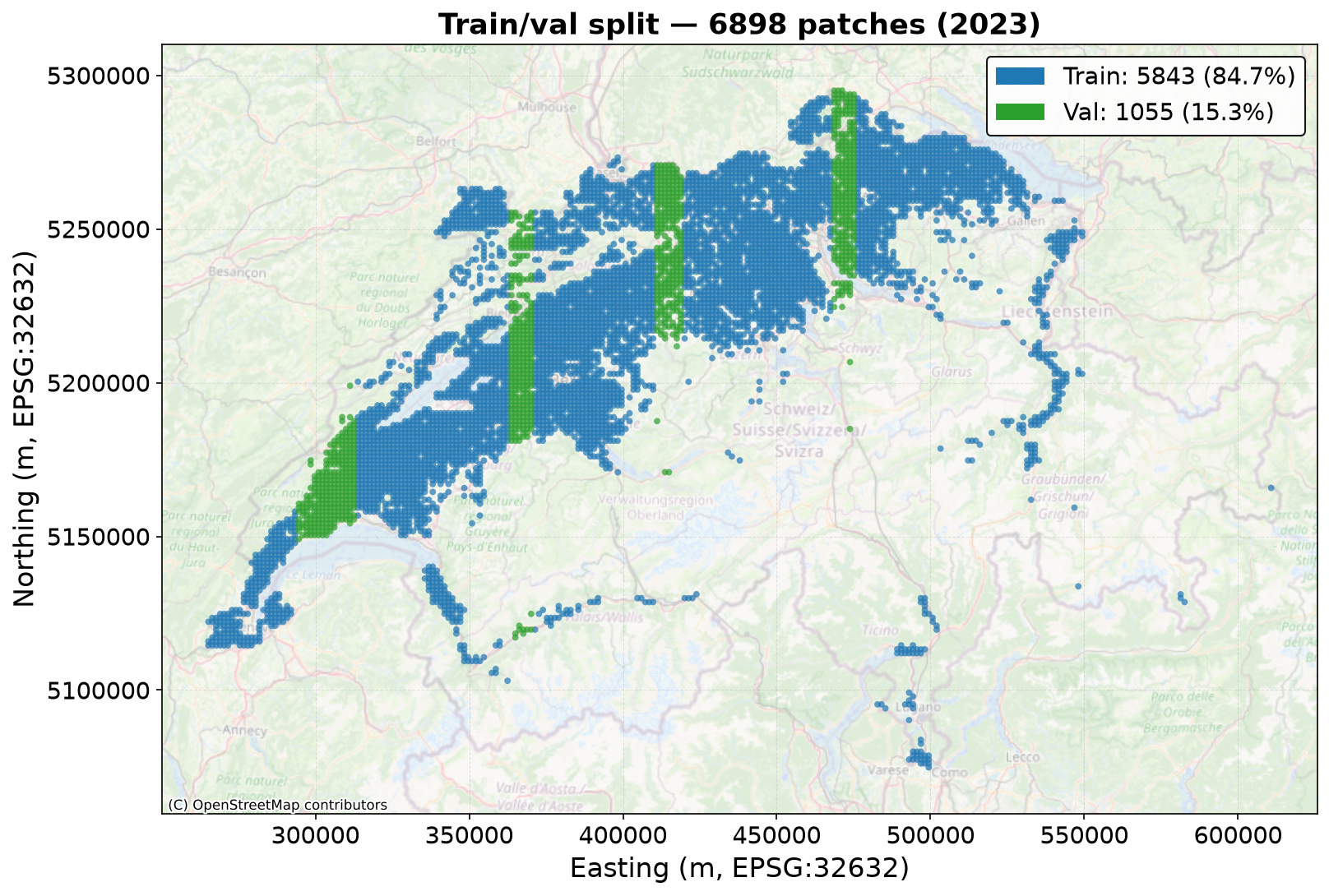}
  \caption{\textbf{Spatial train/validation split over the Swiss winter wheat
  fields.} Each point is a $1.28\times1.28\,\mathrm{km}$ patch (2023 season shown),
  coloured by split over an OpenStreetMap basemap (coordinates in UTM zone~32N,
  EPSG:32632). The validation stripes (green, $\sim$$15\%$) are interleaved four
  times west to east. All remaining patches are used for training (blue).}
  \label{fig:spatial-split}
\end{figure}

App.~\ref{fig:spatial-split} shows the spatial train/validation split of Sec.~\ref{subsec:dataset}: the held-out validation stripes (about $15\%$ of the patches) are interleaved four times by easting across the country, spanning the full west--east climatic gradient, while the remaining patches form the training set.
The split unit is the whole $1.28\times1.28\,\mathrm{km}$ patch.

\section{Training Details and Model Capacity Selection}
\label{sec:appendix-capacity}

The transformer is trained with AdamW under a warmup--cosine schedule and the GRU with Adam at a constant rate.
Both GRU and transformer use a batch size of $512$, a $30$-epoch cap with early stopping, and are trained data-parallel across four GPUs.

\begin{figure}[H]
  \centering
  \includegraphics[width=0.49\linewidth]{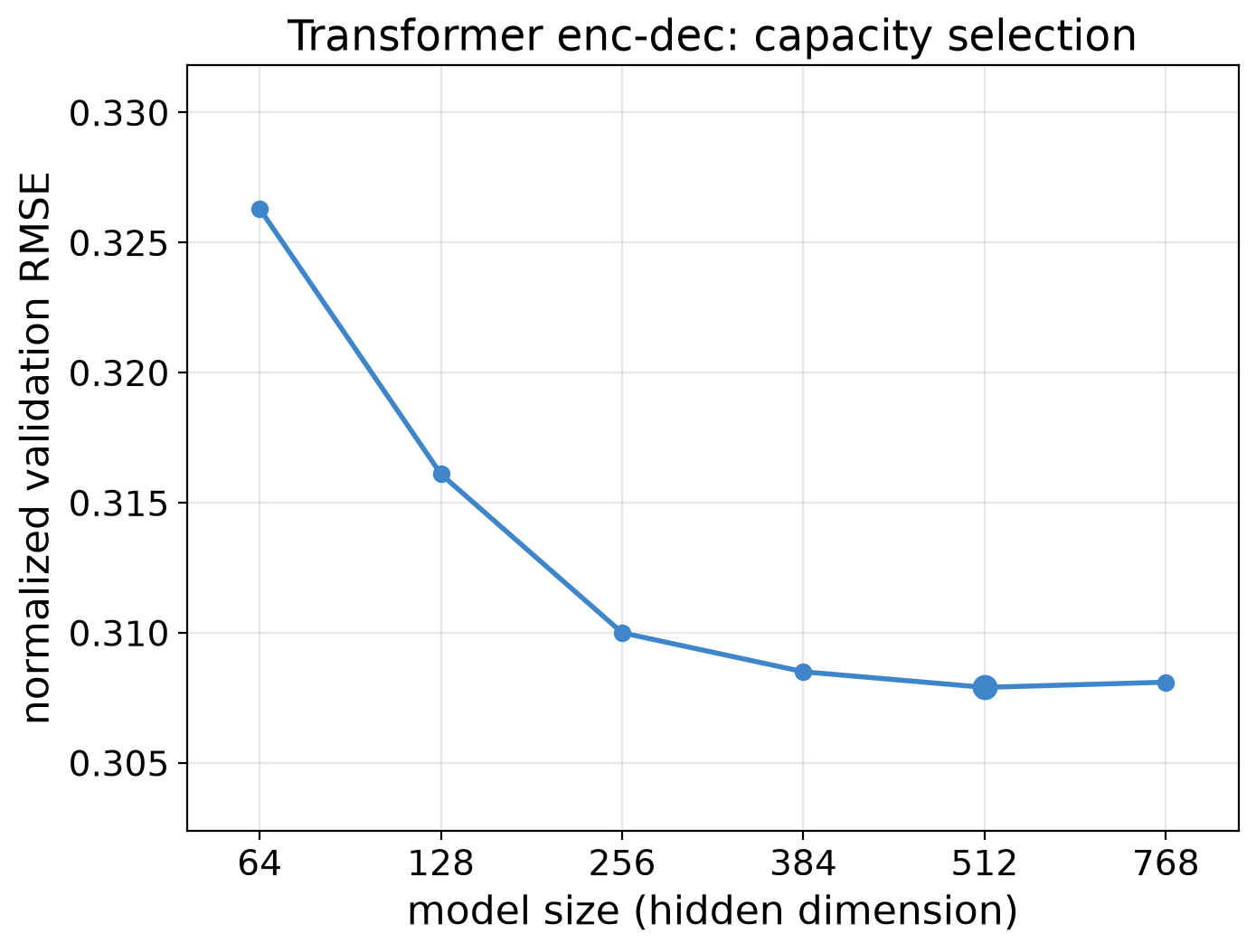}
  \hfill
  \includegraphics[width=0.49\linewidth]{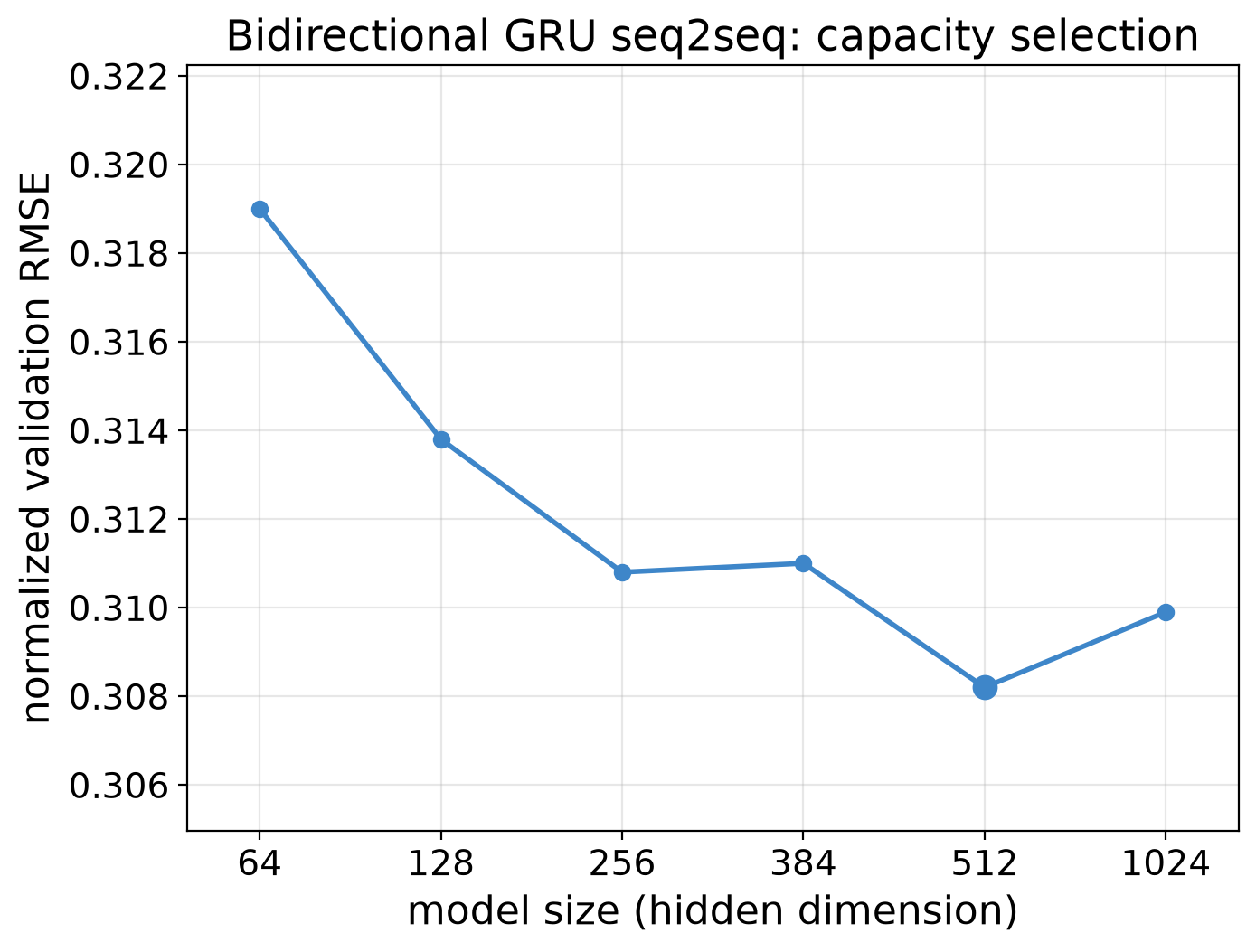}
  \caption{Capacity sweep. Normalised validation RMSE versus model size for each
  family. Accuracy improves with size and then saturates, with
  hidden size $512$ (marked), which is selected for both deep families.}
  \label{fig:capacity}
\end{figure}

We select the model size of each family by a capacity sweep, keeping the size with the best validation loss.
The transformer grows its width and depth together, from a hidden size of $64$ up to $768$ (feed-forward width $4\times$ the model dimension, head dimension $64$, encoder and decoder layers grown together).
The GRU grows its hidden width from $64$ to $1024$ at two layers.
The MLP uses three hidden-layer widths, and LightGBM varies its leaf budget over $L\in\{127,255,511\}$ leaves per tree, where larger values were not considered because of diminishing returns.
For both the transformer and the GRU, validation accuracy rises with size and then levels off, with optimal performance at hidden size $512$ (bigger models stop training earlier and gain no accuracy), so we select hidden size $512$ for both and carry it forward (App.~\ref{fig:capacity}).

\section{Growth--Temperature Response and $\lambda$ Selection}
\label{sec:appendix-thermal}

The growth--temperature selection criterion of Sec.~\ref{subsec:validate-prior} is applied identically to both families; the peak correlation strength against $\lambda$ for the transformer and the GRU is compared in the main text (Fig.~\ref{fig:thermal-lambda}).
Here we show the full season-averaged growth--temperature response curves at the selected $\lambda$ for both families.
Both peak at a 2-day lag, the short delay expected between warming and canopy growth, and both are evaluated on the validation subset over forecast start days DOY~60--90 (predicted days to DOY~122 at the $32$-day horizon).

\begin{figure}[H]
  \centering
  \includegraphics[width=0.49\linewidth]{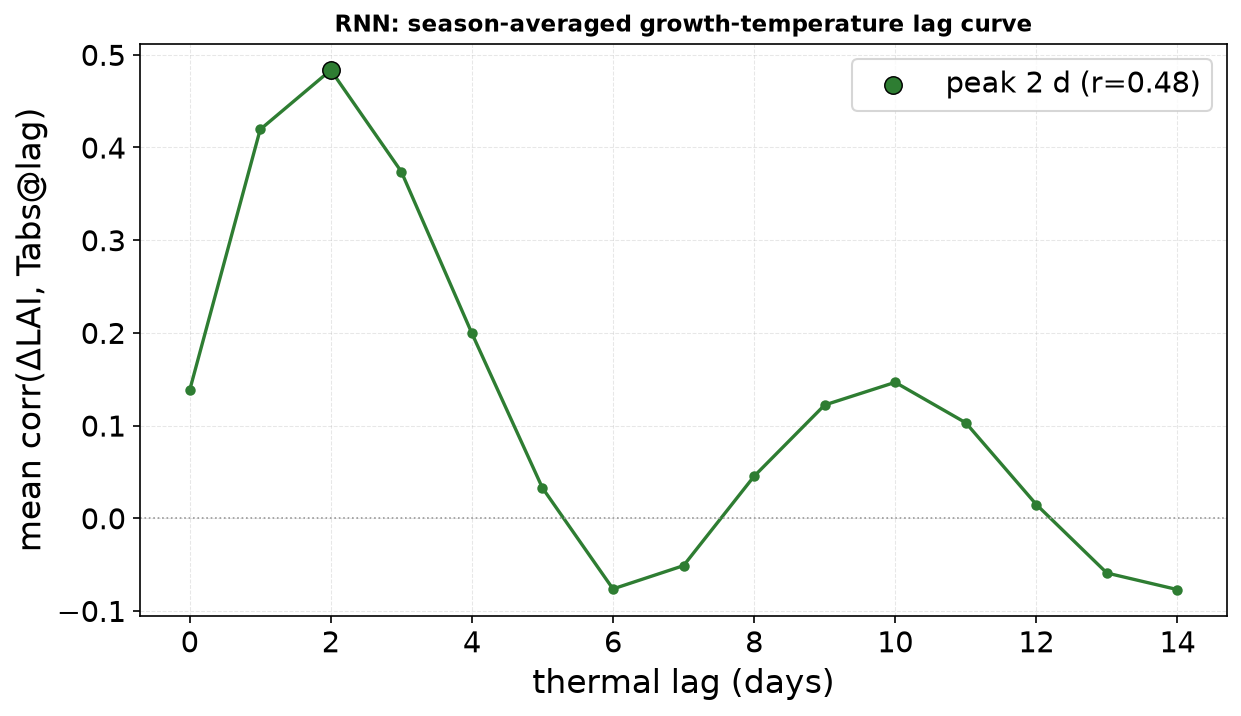}
  \hfill
  \includegraphics[width=0.49\linewidth]{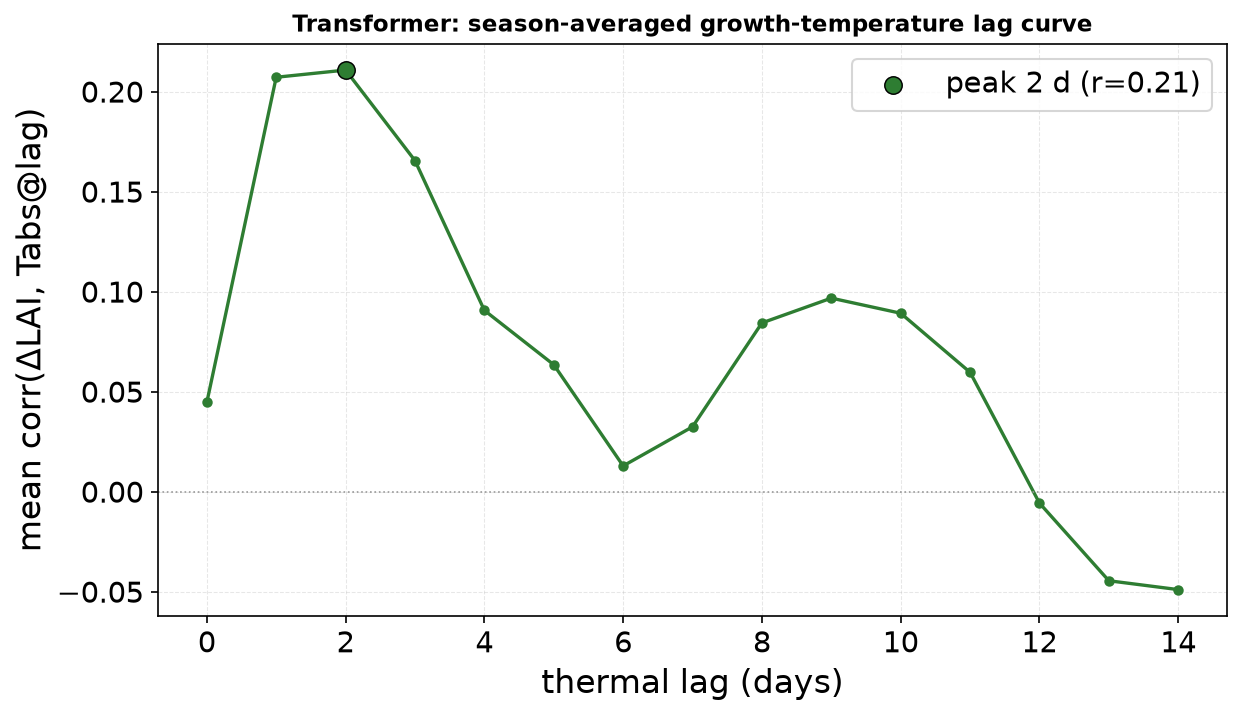}
  \caption{\textbf{Season-averaged growth--temperature response at the selected
  $\lambda$.} Correlation between predicted day-to-day growth ($\Delta$LAI) and
  daily mean temperature as a function of the temperature lead (lag), averaged over
  forecast start days (DOY~60--90, validation subset), for the GRU at
  $\lambda = 0.1$ (left) and the transformer at $\lambda = 0.05$ (right). Both
  responses peak at a 2-day lag (marked), with the 1-day lag almost as strongly
  correlated, matching the short physiological delay
  between warming and canopy growth. The weak secondary structure beyond
  $\sim$8 days reflects residual temperature autocorrelation rather than a direct
  growth response.}
  \label{fig:thermal-curves}
\end{figure}

\section{Soil Moisture as a Non-Observable Driver}
\label{sec:appendix-soil}

Soil moisture is a key driver of crop growth, yet we do not supply it as an input to the model (Sec.~\ref{sec:method}).
Soil moisture varies greatly within the landscape~\cite{zhu_2011_InfluencesSoilTerrain}, but area-wide approaches to monitor or model it only provide coarse spatial resolution ($>$approx.\ $12\,\mathrm{km}$).
Higher-resolution products (e.g., from Sentinel-1) suffer from vegetation cover~\cite{o._2021_GlobalSoilMoisture,mohr_nasa-usda_2024,peng_2021_RoadmapHighresolutionSatellite}.
Consequently, we currently regard soil moisture information as unavailable.
Still, since soil moisture is largely a function of evapotranspiration---which depends largely on LAI and meteorological conditions---we assume that a sufficiently capable model will inherently factor in the water available to the plant from the weather information and the transpirative surface (LAI).
This is why the run-in window is chosen long enough for the model to sense the crop state and approximate the effect of such non-observable drivers.

\section{2023 NRMSE}
\label{sec:appendix-greenup}

\begin{figure}[H]
  \centering
  \includegraphics[width=0.8\linewidth]{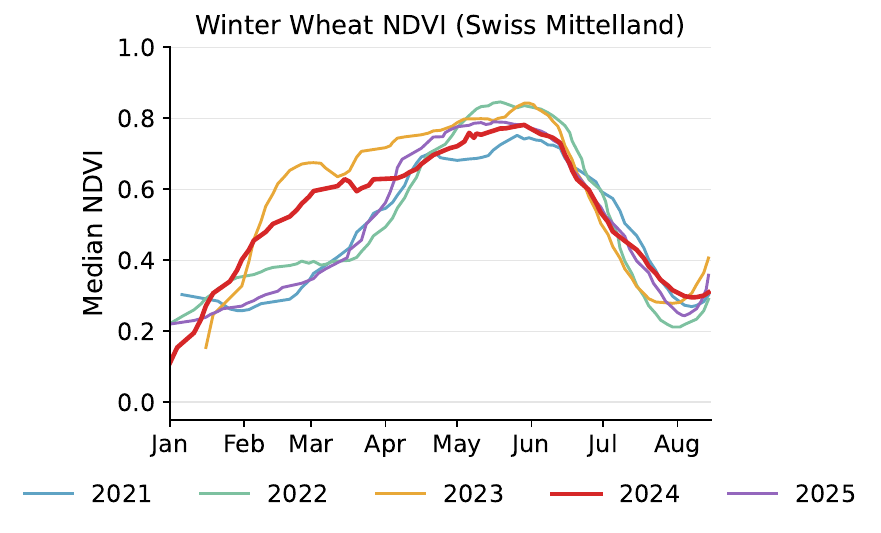}
  \caption{\textbf{Seasonal NDVI of Swiss winter wheat, 2021--2025.} (Figure reproduced from Lauber et al. \cite{lauber2026swisscrop}).
  Median
  NDVI over the winter-wheat area of the Swiss Mittelland as a function of
  calendar date, one curve per season, used as an observable proxy for canopy
  density (LAI). The 2023 season (orange) rises earlier and more steeply through
  late winter and spring and remains above the other seasons until early summer,
  so its observed canopy is on average denser. This raises the mean observed LAI
  that NRMSE normalises by Eq.~\eqref{eq:nrmse} and explains the anomalously low
  2023 NRMSE in Table~\ref{tab:per-year}.}
  \label{fig:ndvi-greenup}
\end{figure}

For the GRU, the per-year metrics of Table~\ref{tab:per-year} rank the five held-out seasons consistently under RMSE, MAE and $\text{R}^2$, but 2023 is an outlier under NRMSE: it records the lowest NRMSE of any fold despite a high RMSE.
Since NRMSE normalises the RMSE by the mean observed LAI (Eq.~\eqref{eq:nrmse}), a season whose observed canopy is on average denser is scored favourably.
App.~\ref{fig:ndvi-greenup} shows why 2023 is such a season: using median NDVI over the winter-wheat area of the Swiss Mittelland as an observable proxy for canopy density, 2023 greens up earlier and faster than the other four seasons and stays above them through spring.
The crop therefore spent comparatively little of the observed window at low leaf area, so there are few low-LAI samples, the mean observed LAI $\bar{y}$ is high, and the NRMSE---the RMSE divided by that mean---is correspondingly low even though the absolute error is not.

\section{Forecast Error Across the Prediction Horizon}
\label{sec:appendix-horizon}

\begin{figure}[H]
  \centering
  \includegraphics[width=0.7\linewidth]{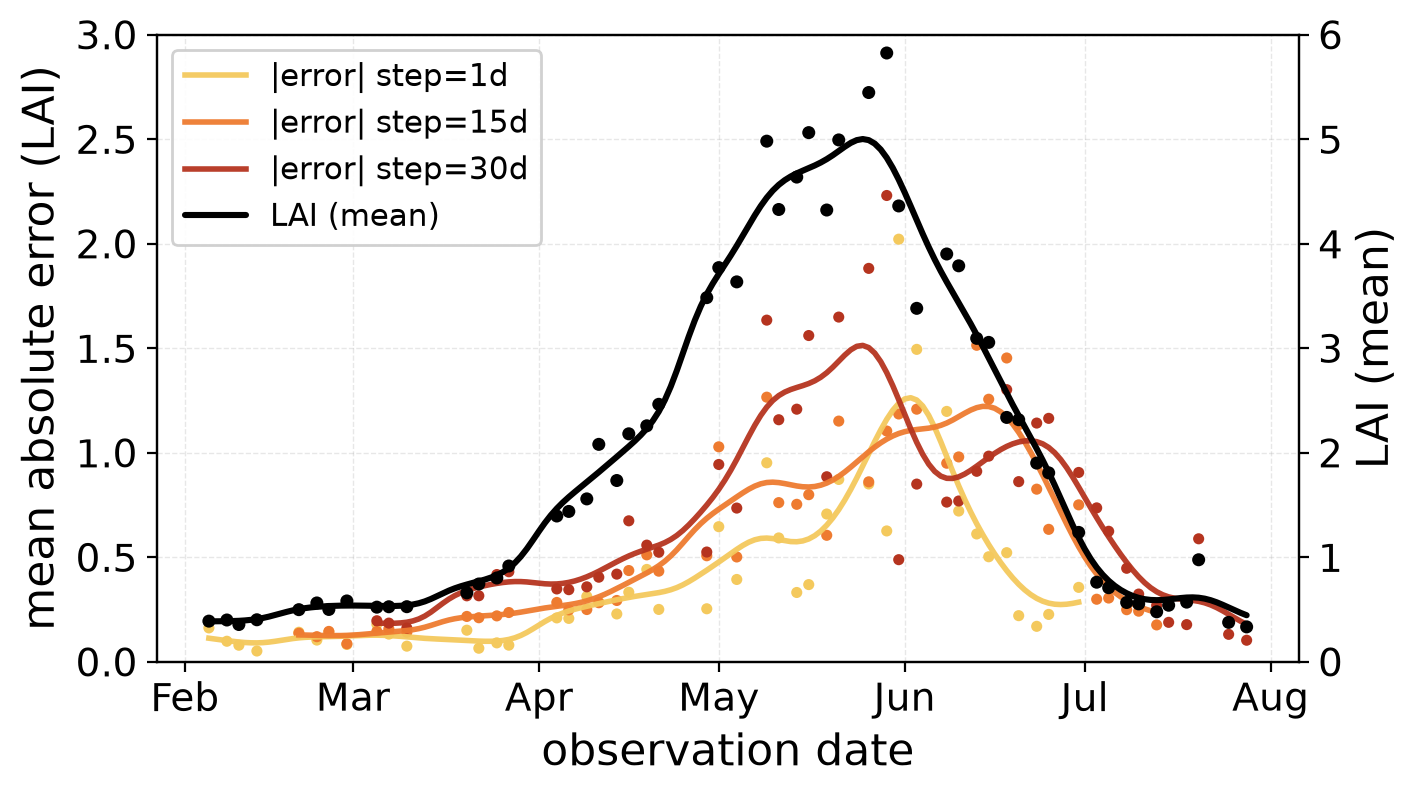}
  \caption{\textbf{Transformer forecast error across the prediction horizon}
  (Z\"urich, 2022 held-out season). Mean absolute forecast error at $1$-, $15$- and
  $30$-day lead times (left axis) averaged over the hundred patches nearest the
  Z\"urich city centre, with their mean ground-truth LAI (black, right axis), each
  smoothed with a Gaussian kernel ($\sigma=5$ days). Error grows only modestly from
  the $1$-day to the $30$-day forecast, mirroring the GRU
  (Fig.~\ref{fig:loyo-abserr-horizon}).}
  \label{fig:loyo-abserr-horizon-tfm}
\end{figure}

App.~\ref{fig:loyo-abserr-horizon-tfm} repeats the lead-time analysis of Sec.~\ref{subsec:main-results} for the transformer. As in the main text, the curves are computed over the hundred $1.28\times1.28\,\mathrm{km}$ patches containing winter wheat nearest the Z\"urich city centre in the 2022 held-out season: for every Sentinel-2 observation date we average the absolute forecast error at $1$-, $15$- and $30$-day lead times over all valid wheat pixels of those patches, alongside their mean ground-truth LAI, and smooth each series with a Gaussian kernel ($\sigma = 5$ days). The transformer shows the same gentle growth of error with lead time, and the same seasonal envelope, as the GRU in the main text.

\section{Evaluation Metrics}
\label{sec:appendix-metrics}

All accuracy numbers in this paper (Table~\ref{tab:per-year}, Fig.~\ref{fig:model-selection}, Fig.~\ref{fig:lambda} etc.) are computed with a single, shared evaluator so that every model, fold, and split is scored identically.
This appendix gives the exact definitions.

\paragraph{\textbf{Evaluation set and masking.}}
A forecast produces one predicted LAI value $\hat{y}$ per (pixel, day) inside each prediction window, over horizons $h\in[1,32]$ days.
Let the evaluation set be the pool of all such (pixel, day) pairs over every window, patch, and forecast start day in the fold being scored.
A pair is \emph{valid} only if both the ground-truth LAI $y$ and the prediction $\hat{y}$ are finite; pairs with a missing (NaN) ground truth or prediction are dropped before any metric is formed.
Write the $N$ valid pairs as $\{(y_i,\hat{y}_i)\}_{i=1}^{N}$ and the mean observed LAI as
\begin{equation}
  \bar{y} \;=\; \frac{1}{N}\sum_{i=1}^{N} y_i .
  \label{eq:mean-gt}
\end{equation}
Every metric below is accumulated in a single streaming pass over the fold's prediction shards (only running sums of $y_i$, $y_i^2$, $\hat{y}_i$, the squared error, and the absolute error are kept), so pooling over pixels, days, horizons, and patches is exact and never materialises the full residual vector.
For the country-scale folds of Table~\ref{tab:per-year} the accumulators are summed over the entire held-out season before any metric is formed, so each reported number is an exact sample-count--weighted aggregate over all valid pairs rather than an average of separately scored subsets.

\paragraph{\textbf{Root-mean-square error (RMSE, LAI units, $\downarrow$).}}
The square root of the mean squared forecast error,
\begin{equation}
  \mathrm{RMSE} \;=\; \sqrt{\frac{1}{N}\sum_{i=1}^{N}\bigl(y_i-\hat{y}_i\bigr)^2}\,.
  \label{eq:rmse}
\end{equation}
It is in the same units as LAI and, through the square, penalises large errors more strongly than small ones.

\paragraph{\textbf{Normalised RMSE (NRMSE, \%, $\downarrow$).}}
Absolute LAI error is not directly comparable across seasons, because canopies differ in overall greenness: the same $0.7$ LAI RMSE is a mild error over a dense summer canopy but a severe one over a sparse one.
We therefore report a scale-free companion to the RMSE, obtained by normalising it by the mean observed LAI of the evaluation set,
\begin{equation}
  \mathrm{NRMSE} \;=\; \frac{\mathrm{RMSE}}{\bar{y}}\,\times\,100\,\%,
  \label{eq:nrmse}
\end{equation}
with the RMSE of Eq.~\eqref{eq:rmse} in the numerator and the mean observed LAI $\bar{y}$ of Eq.~\eqref{eq:mean-gt} in the denominator, i.e.\ the same $N$ valid (pixel, day) pairs enter both.
NRMSE is dimensionless and expresses the typical forecast error as a fraction of how much leaf area there is to predict, so it is comparable between folds: a season whose canopy is on average sparser (smaller $\bar{y}$) yields a larger NRMSE for the same absolute RMSE, and a $10\,\%$ NRMSE means the root-mean-square error is one tenth of the average observed LAI.
We normalise by the \emph{mean} of $y$ rather than by its range, so that the statistic is governed by the bulk of the canopy rather than by a handful of extreme LAI values; and we prefer it to a per-pixel mean absolute percentage error, which divides each residual by its own $y_i$ and is dominated by the many pixels with near-zero LAI, where a small absolute error becomes an enormous relative one.

\paragraph{\textbf{Mean absolute error (MAE, LAI units, $\downarrow$).}}
The mean of the absolute forecast errors,
\begin{equation}
  \mathrm{MAE} \;=\; \frac{1}{N}\sum_{i=1}^{N}\bigl|y_i-\hat{y}_i\bigr|\,.
  \label{eq:mae}
\end{equation}
Also in LAI units, it weights every error linearly and is therefore less sensitive to a few large residuals than RMSE; the gap $\mathrm{RMSE}\!-\!\mathrm{MAE}$ reflects the spread of the error distribution.

\paragraph{\textbf{Coefficient of determination ($\text{R}^2$, dimensionless, $\uparrow$).}}
The fraction of the observed LAI variance explained by the forecast,
\begin{equation}
  \text{R}^2 \;=\; 1 \;-\; \frac{\sum_{i=1}^{N}\bigl(y_i-\hat{y}_i\bigr)^2}{\sum_{i=1}^{N}\bigl(y_i-\bar{y}\bigr)^2}\,,
  \label{eq:r2}
\end{equation}
comparing the model against the constant predictor $\hat{y}_i\equiv\bar{y}$: $\text{R}^2=1$ is a perfect fit, $\text{R}^2=0$ matches the mean predictor, and negative values are worse than it.
The denominator is the total sum of squares of the same $N$ valid pairs; a $10^{-8}$ term is added to it in the implementation purely to guard against division by zero and has no effect at the sample sizes used here.

\end{document}